# What do we learn about development from baby robots?[1]

Pierre-Yves Oudeyer
Inria and EnstaParisTech, France
http://www.pyoudeyer.com

To understand the world around them, human beings fabricate and experiment. Children endlessly build, destroy, and manipulate to make sense of objects, forces, and people. To understand boats and water,for example, they throw wooden sticks into rivers. Jean Piaget, one of the pioneers of developmental psychology, extensively studied this central role of action in infant learning and discovery (Piaget, 1952).

Scientists do the same thing: To understand ocean waves, we build giant aquariums. To understand cells, we break them down to their component parts. To understand the formation of spiral galaxies, wemanipulate them in computer simulations. Constructing artifacts helps us construct knowledge.

But what if we want to understand *ourselves*? How can we understand the mechanisms of human learning, emotions,and curiosity? Here, the physical fabrication of artifacts can also be useful. Researchers can actually build *baby robots* with mechanisms that model aspects of the infant brain and body, and  thenalter these models systematically (see figure 1). We can compare the behavior we observe with the mechanisms inside. Indeed, robots are now becoming an essential tool to explore the complexity of development, a tool that allows scientists to grasp the complicated dynamics of a child's mind and behavior.

## EXPLORING COMPLEX SYSTEMS

Modern developmental science has now invalidated the old divide between nature and nurture. We now know that genes are not a static program that unfolds independently of the environment. We also understand that learning in the real world can only work if there are appropriate constraints during development (Gottlieb, 1991; Lickliter& Honeycutt, 2009). Finally, we know that many behavioral and cognitive patterns cannot be explained by reducing them to single genes, organs, or isolated features of the environment: they result from the *dynamic interaction* between cells, organs, learning mechanisms, and the physical and social properties of the environment at multiple spatiotemporal scales. Development is a complex dynamical system, characterized by the spontaneous self-organization of patterns, sometimes called "emergent patterns" (Thelen and Smith, 1996).

The concepts of complex systems and self-organization revolutionized physics in the 20[th] century. They characterize phenomena as diverse as the formation of ice crystals, sand dunes, water bubbles, climatic structures, and galaxies. A key to these scientific advances was the use of mathematics and computer simulations. Indeed, it is hard to imagine how we could understand the dynamics of ice crystals or the formation of clouds without mathematics and computer simulations. At the end of the 20[th] century, biologists began to use these concepts, for example, to understand the formation of termite nests (Ball, 2001, see figure 2). They used computer simulations to show how the local interaction between thousands of little termites, with no plan of the global structure, could self-organize sophisticated and functional large-scale architectures. Other mathematical and computational models were similarly used to study the self-organization of stripes and spots on the skin of animals, spiral of horns and mollusk shells, patterns of the dynamics of predator-prey populations or of the dynamics of heart beat (Ball, 2001).

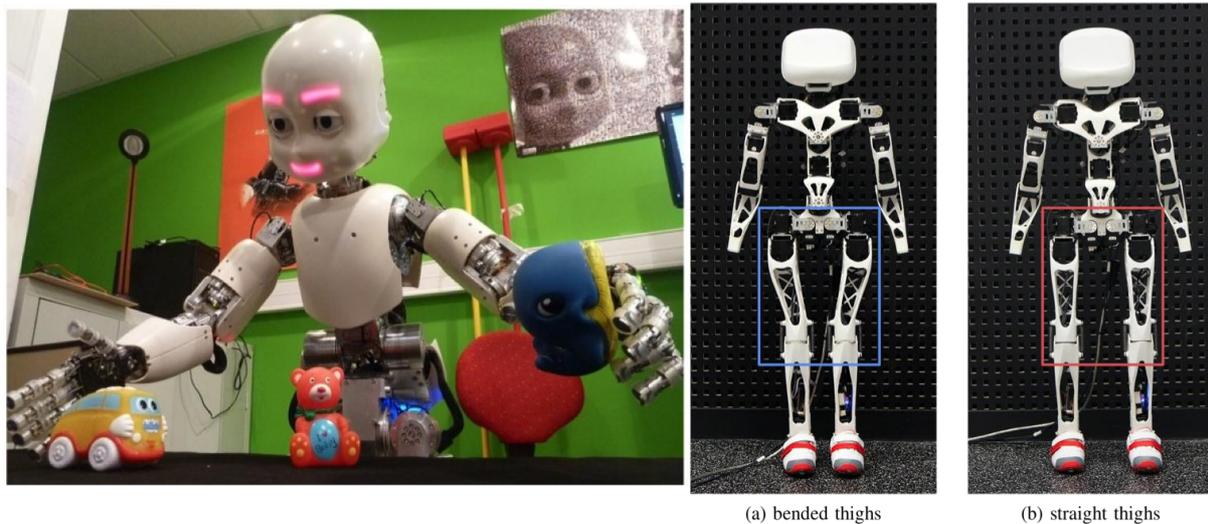

(a) bended thighs       (b) straight thighs

**Figure 1Open-Source Baby Robots.** Robots can help us model and study the complex interaction between the brain, the body and the environment during cognitive development. Here we see two open-source robotic platforms used in laboratories. Being open-source allows open science through revealing all details in the experiments as well as replicability. Based on 3D printing, the Poppy platform allows fast and efficient exploration of various body morphologies (Lapeyre et al., 2014), such as leg shape (see alternatives on the right (a) and (b)), and how this can impact development of skills. Left: ICub http://www.icub.org , right: Poppy http://www.poppy-project.org.

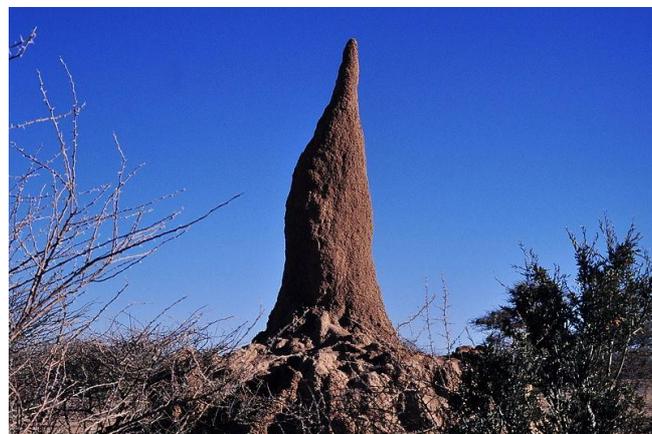

**Figure 2Termite Nest.** The architecture of termite nests is the self-organized result of the interaction of thousands of insects, but none of these insects has a map of the architecture. Computer simulations contributed to understanding this process.Photo: http://commons.wikimedia.org/wiki/File:Termite%27s_nest.jpg

Child development also involves the interaction of many components, but in a way that is probably orders of magnitude more complicated than crystal formation or termite nest construction. Hence, to complement the (tremendously useful) verbal conceptual tools of psychology and biology, researchers have begun building machines that model pattern formation in development. Such efforts can play a key role in 21st century developmental science.

Building machines that learn and develop like infants is actually not a new idea. Alan Turing, who helped invent the first computers in the 1940s, already had the intuition that machines could be useful to understand processes of psychology :

> *"Instead of trying to produce a programme to simulate the adult mind, why not rather try to produce one which simulates the child's? If this were then subjected to an appropriate course of education one would obtain the adult brain. Presumably the child brain is something like a notebook as one buys it from the stationer's. Rather little mechanism, and lots of blank sheets. "*
> *(Turing, 1950)*

For several reasons, Turing's vision was not transformed into a concrete scientific program until the very end of the 20th century. First, the 1950s saw the rise of cognitivism and artificial intelligence, views which promoted the (unsuccessful) idea that intelligence could be seen as an abstract symbol manipulation system that could be handcrafted directly in its adult form. Second, Turing missed two important elements:A)Learning from a blank slate-a tabula rasa—could not work in real organisms facing the complex flow of information and action in the world. Rather, development needsconstraints. B) Turing missed the role of the body: behavior and cognition arise in a physical substrate, and this physical substrate strongly influences development (Pfeifer and Bongard, 2007). The key role of the body is the reason why robots, and not simply abstract computer simulations, can be key in developmental science.

**THE ROLE OF THE BODY IN BIPED WALKING**

Let us look at some examples, beginning with the behavior of biped walking. While this is a very familiar skill, we are nevertheless far from understanding how we walk with two legs, and how infants learn to do this. What is walking? What does it mean to acquire the capability to walk on two legs? Walking implies the real-time coordination of many body parts. Each of our bones and each of our muscles are like the musicians of a symphonic orchestra: they must produce a movement impulse (or silence) at the right moment; and it is the juxtaposition and integration of all these impulses and silences which builds the symphony of the whole body walking forward with elegance and robustness.

But is there a musical score which plans these coordination details? Is there a conductor driving the movement? In technical terms, is walking equivalent to calculating? Does the brain, every few milliseconds, observe the current state of the body and environment and compute the right muscular activations to maintain balance and move forward with minimal energy consumption?

Viewing walking as pure computation is the approach which has long been taken by specialists who study human walking. Viewed in this context, understanding the development of walking requires understanding how the child could develop the capability to achieve all these real-time computations and make predictions about the dynamics of its body. Some roboticists interested in having biped robots walk also tried this approach.Yet, even if sometimes there are beautiful performances (Hirose and Ogawa, 2007), this has so far led to humanoids that fall very easily, with a very unnatural walking style.

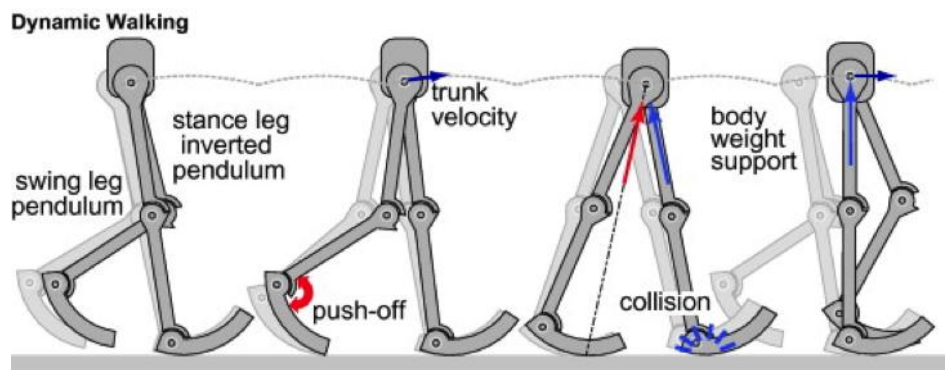

**Figure 3Passive dynamic walker robot.** A robot that walks but does not have a brain ! It has neither electric power nor computer. Through its shape, its steps are spontaneously generated through the physical interaction between its structure and gravity (Adapted from http://dyros.snu.ac.kr/concept-of-passive-dynamic-walking-robot/).

Perhaps, then, walking is much more than calculation.Twenty years ago, a genius roboticistnamed Tad McGeerconducted an experiment that changed our understanding of biped walking in humans and machines. He built a pair of mechanical legs (see Figure 3 and video https://www.youtube.com/watch?v=WOPED7I5Lac), without a motor and without a computer (thus without the possibility to make calculations), and reproduced the geometry of human legs (McGeer, 1990).Then, he threw the robot on a little slope, and the robot walked: automatically, through the physical interaction between the various mechanical parts and gravity, the two legs generated a gait that looked surprisingly similar to a human gait, and was robust to disturbances. Other laboratories replicated the experiment many times (e.g. see video https://www.youtube.com/watch?v=rhu2xNIpgDE) and showed that this biped movement could last forever on a treadmill (Collins et al., 2005). The coordination of the set of mechanical parts of this robot, interacting only locally through their physical contacts, is *self-organized*: there is no predefined plan for the coordination, and no conductor synchronizing every part of the score. Walking is a dynamic emergent pattern where physics and the body have a fundamental role, each providing structure and constraints for the other.

These very clever experiments show how robots can be used to disentangle the roles of the body and the neural system within a model of walking. And at the same time, they articulate experimentally concepts that enlight our understanding of human development. Here, we observe the self-organization of a pattern (biped walking) that is neither innate (there are no genes and no program) nor learned (there is no learning taking place). This concretely demonstrates that the divide between innateness and learning can be meaningless. Structure can appear spontaneously through complex biophysical interactions. And such structure can then be leveraged for learning and development. Such an experiment allowed to formulate and ground solidly the hypothesis that learning how to walk may boil down to learning to reuse and tune a movement structure already embedded in the dynamics of the body.

## SELF-ORGANIZATION OF CURIOSITY-DRIVEN DEVELOPMENTAL PROCESSES

Let us examine a second example that concerns the role of curiosity in child development. Human children learn many things, often in a progressive way with a specific timing and ordering. For example, before they learn to walk with their two legs, they first explore how to control their neck, then to roll on their belly, then to sit, to stand up, and walk with their hands on the walls. Why and how do they follow this particular progression? Also, many steps of developmental trajectories appear in a similar ordering in many children, but at the same time some children follow quite different developmental paths. How can we explain the apparent universal tendencies on one hand and the individual variability on the other? Is universality the result of a "program"? And when we observe diverging developmental paths, does this mean that something in the "program" is necessarily broken?

The social environment plays a big role in guiding developmental process, and has been the object of study of robot modeling work focusing on the roles of *imitation* (Breazeal and Scassellati, 2002; Nehaniv and Dautenhahn, 2007; Demiris and Meltzoff, 2008), *joint attention* (Nagai et al., 2006), *language* (Steels, 2012; Cangelosi et al., 2010), and *interactive alignment of tutor and learner* (Vollmer et al., 2014). But there is another fundamental force which drives all of us: *curiosity*, which pushes us to discover, to create, to invent.

Research in psychology and neuroscience has identified that our brains have an intrinsic motivation to explore novel activities for the sake of learning and practicing (Lowenstein, 1994). Yet we still understand little about curiosity and how it impacts development. Neuroscientists are only beginning to identify brain circuits involved in spontaneous exploratory behaviors (Gottlieb et al., 2013).

Several research teams have proposed to advance our understanding of curiosity and its impact on development by fabricating robots that learn, discover, and generate their own goals with models of curiosity-driven learning (Baldassarre and Mirolli, 2013; Gottlieb et al., 2013; Oudeyer and Kaplan, 2007). An example comes from the Playground experiment (see Figure 4 and video https://www.youtube.com/watch?v=uAoNzHjzzys ;Oudeyer et al., 2007; Oudeyer and Smith, 2014). Here, a robot learns by making experiments: he tries actions, observes effects, and detects regularities between these actions and their effects. This allows him to make predictions. The way he chooses actions is like a little scientist: he chooses experiments which he thinks can improve his own predictions, which can provide new information, which make it progress in learning, while continuously allocating some proportion of time to exploring other activities in search of new potential niches of progress. The robot is also equipped with a mechanism that simultaneously categorizes its sensorimotor experiences into different categoriesbased on how similar they are in learnability and controllability.

At any moment in its development, the robot mainly focuses on exploring activities that are sources of learning progress, those that are neither too easy nor too difficult. This models the idea that what the brain finds interesting to practice is what is just beyond the current level of knowledge or competencies (Csikszenthmihalyi, 1991; Schmidhuber, 1991; Kidd, Piantadosi, &Aslin, 2012). Such a model also leads to a concrete definition of curiosity as a motivational mechanism that pushes an organism to explore activities for the primary sake of gaining information (as opposed to searching for information to achieve an external goal like finding food or shelter).

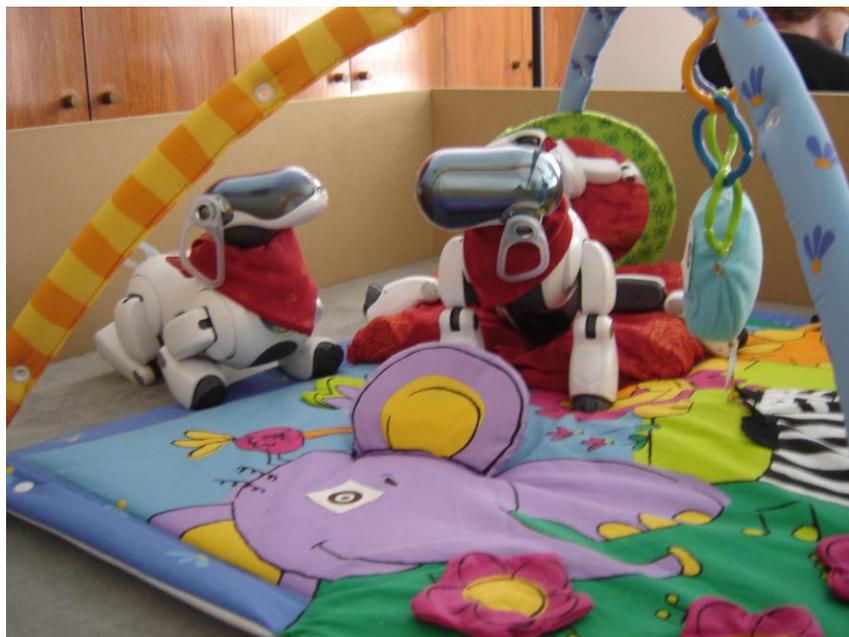

**Figure 4Curiosity-driven learning in the Playground Experiment.** The Playground Experiment (Oudeyer et al., 2007 ; Oudeyer and Smith, 2014). The robot in the center explores and learns to predict the effects of its actions, driven by a form of artificial curiosity. The behaviors and know-how it acquires spontaneously evolve and self-organize into developmental stages of increasing complexity, without an initial program specifying these stages. For example, the robot learns at its own initiative how to grasp an object in front of him, or how to produce vocalizations which provoke reactions in the other robot.

In the Playground experiment, one observes that not only is the robot able to learn skills based on its own initiative—for example, by learning how to grasp the object in front of it—but also to spontaneously evolve and self-organize its behavior, progressively increasing in complexity. Cognitive stages appear, but they are not pre-programmed. For example, after beginning through relatively random body babbling, the robot often focuses first on moving the legs around to predict how it can touch objects, then focused on grasping an object with its mouth, and finally ends up exploring vocal interaction with the other robot. Critically, the engineer preprogrammed neither of these specific activities, nor did the engineer preprogram their timing and ordering.

This self-organization results from the dynamic interaction between curiosity, learning, and the properties of the body and environment.If the same experiment is repeated several times with the same parameters, one observes that often the same coarse developmental stages appear. Yet, sometimes individuals invert stages, or even generate qualitatively different behaviors. This is due to random small contingencies, to even small variability in the physical realities, and to the fact that this developmental dynamic system has what mathematicians call "attractors": a collection of differentiated states towards which the system spontaneously evolves as soon as it finds itself in their vicinity, called "basins of attraction".

This robot experiment helps us to understand and formulate hypotheses about how development works. It suggests a way to model the mechanisms of curiosity-driven learning, and to assess how curiosity can be modeled as a concrete mechanism within a physical agent. It also shows how in the long-term curiosity-driven developmental process can self-organize developmental stages of increasing complexity, without a predefined maturational schedule. Finally, it offers a way to understand individual differences as emergent in development, making clear how developmental process might vary across contexts, even with an identical underlying mechanism.

## SUMMARY

Understanding infant development is one of the greatest scientific challenges of contemporary science, and we are only beginning to uncover its basic mechanisms. A large source of difficulty comes from the fact that the development of skills in infant results from the interactions of multiple mechanisms at multiple spatial (molecules, genes, cells, organs, bodies, social groups) and temporal scales. Like spiral galaxies which shape is neither the programmatic unfolding of a plan nor the result of learning, infant development is pervaded with patterns that form spontaneously out of a complex distributed network of forces. The concepts of "innate" or "acquired" are not any more adequate tools for explanations, which call for a shift from reductionist to systemic accounts.

As physics realized a long time ago, systemic explanations of pattern formation in complex systems require the use of formal models based on mathematics and algorithms, which allow us to fabricate and simulate aspects of reality. In the words of Nobel prize physicist Richard Feynman:

*"What I cannot create I cannot understand"*

Such an approach is now being taken to developmental science, where algorithmic and robotic models are used to explore the dynamics of pattern formation in sensorimotor, cognitive and social development. Formulating hypothesis about development using such models, and exploring them through experiments with simulation and robots, allows us to consider the interaction between many mechanisms and parameters. This crucially complements traditional experimental methods in psychology and neuroscience where only a few variables can be studied at the same time.

Furthermore, the use of robots is of particular importance. The laws of physics generate everywhere around us spontaneous patterns in the inorganic world (ice crystals, clouds, dunes, river deltas, …). They also strongly impact the living, and in particular constrain and guide infant development through the properties of its (changing) body in interaction with the physical environment. Being able to consider the body as an experimental variable, something that can be systematically changed in order to study the impact on skill formation, has been a dream to many developmental scientists. This is today becoming possible with robotics (Kaplan and Oudeyer, 2009).